\pdfoutput=1

\documentclass[11pt]{article}

\usepackage{submission}
\usepackage{epsfig}
\usepackage{amssymb}

\usepackage{caption} 
\usepackage{amsmath,graphicx}
\newcommand{\plus}{\scalebox{0.6}{$+$}}
\usepackage{longtable}
\usepackage{xcolor}
  
\usepackage{times}
\usepackage{latexsym}

\usepackage{url}
\usepackage{hyperref}
\usepackage{algpseudocode}

\usepackage{wrapfig}
\usepackage{algorithm}

\usepackage{float}
\usepackage{placeins}

\usepackage{color}

\usepackage[T1]{fontenc}

\usepackage[utf8]{inputenc}

\usepackage{microtype}

\usepackage{subfigure}

\usepackage[
  bottom,
  hang,
  multiple
]{footmisc}
\usepackage{blindtext}

\usepackage{amssymb}
\usepackage{pifont}
\newcommand{\cmark}{\ding{51}}%
\newcommand{\xmark}{\ding{55}}%

\makeatletter
\newcommand{\printfnsymbol}[1]{%
  \textsuperscript{\@fnsymbol{\#1}}%
}
\makeatother

\usepackage{varwidth}
\DeclareCaptionFormat{newformat}{%
  \begin{varwidth}{\linewidth}%
    \centering
    #1#2#3%
  \end{varwidth}%
}
\DeclareUnicodeCharacter{2212}{-}

\begin{document}
\title{TableQuery: Querying tabular data with natural language}
\author{Abhijith Neil Abraham$^*$ \\  Saama  \\ abhijithneilabrahampk@gmail.com
        \and   
        Fariz Rahman$^*$ \\ Activeloop \\ fariz@activeloop.ai
        \and
        Damanpreet Kaur \\ Oregon State University \\ kaurd@oregonstate.edu }
\maketitle
\begin{abstract}
This paper presents TableQuery, a novel tool for querying tabular data using deep learning models pre-trained to answer questions on free text. Existing deep learning methods for question answering on tabular data have various limitations, such as having to feed the entire table as input into a neural network model, making them unsuitable for most real-world applications. Since real-world data might contain millions of rows, it may not entirely fit into the memory. Moreover, data could be stored in live databases, which are updated in real-time, and it is impractical to serialize an entire database to a neural network-friendly format each time it is updated. In TableQuery, we use deep learning models pre-trained for question answering on free text to convert natural language queries to structured queries, which can be run against a database or a spreadsheet. This method eliminates the need for fitting the entire data into memory as well as serializing databases. Furthermore, deep learning models pre-trained for question answering on free text are readily available on platforms such as HuggingFace Model Hub \cite{HuggingFaceModels}. TableQuery does not require re-training; when a newly trained model for question answering with better performance is available, it can replace the existing model in TableQuery.
\end{abstract}

\section{Introduction}
\def\thefootnote{*}\footnotetext{These authors contributed equally to this work}

With most businesses growing digital, they are churning out large amounts of data daily. Most of this data is tabular and is dumped into spreadsheets or database tables. However, it is difficult for users from a non-technical background to analyze such large datasets. 
Thus, a solution to query the database using natural language makes it easier to perform this task.

The early work in querying databases using natural language focused on developing supervised models trained end-to-end like TAPAS \cite{herzig-etal-2020-tapas}, Seq2SQL \cite{zhongSeq2SQL2017}. Although these models perform well, they are not highly generalizable and require re-training when the data domain is changed.

The tabular data in real-world applications can be humongous. It is a challenge to attend over the whole data as these datasets do not fit into the memory. TableQuery eliminates the need to fit the dataset into the memory as it uses structured queries to fetch the desired information from the stored database or spreadsheet.

\section{Related Work}
Querying a database using natural language has been studied extensively in the last few years. Earlier work in this area allows learning semantic parsers from natural language data \cite{tang2000automated, yaghmazadeh2017sqlizer,ge2005statistical}. This field of work used a single database for training and testing and had limitations for the types of structured queries it could predict. Such models did not generalize well to domains different from the ones on which they were trained.

Another set of techniques to approach this problem is using neural sequence-to-sequence models that predict parts of queries sequentially \cite{vinyals2015order, neelakantan2015neural, cho2018adversarial, zhongSeq2SQL2017}. \citealp{xu2017sqlnet} proposed an approach to solving the serialization problem, which was observed in sequence-to-sequence models. They introduced a sequence-to-set model and column attention mechanism to synthesize the query based on the proposed sketch. \citealp{mueller2019answering} encoded tables and queries as graphs and answers are selected using a decoder pointer network. Using pre-trained networks on a different task for improving performance on the target task is commonly used in natural language processing. More recently, \citealp{herzig-etal-2020-tapas} proposed TAPAS (TAble PArSing) which extended BERT's architecture \cite{devlin2018bert} to encode tabular data and pre-trained the network on tables and text segments. TAPAS is the current state-of-the-art in querying tabular data. Another approach introduced by \citealp{couderc2015fr2sql} used a tree tagging approach that filters words according to the assigned POS tags. The main problem with their approach is that they cannot generate a query unless the user uses the column name as such from the table in the query or provides a mapping of the column name with the user-desired input. 
\\
We devised an approach to utilize the existing question answering models trained on free-text to be used for tabular data. This allows us to overcome the limitations of the existing models that require separate model fine-tuning when the data domain is changed. We also allow users to query large datasets without any memory restrictions by storing the tables into a database or spreadsheet, unlike some of the existing works \cite{herzig-etal-2020-tapas} that are not useful in real-world applications due to memory limitations.
\raggedright \let\thefootnote\relax\footnote{\url{https://github.com/abhijithneilabraham/tableQA}.}

\section{System Description}

\begin{figure*}
    \centering
  \includegraphics[width=0.8\textwidth, height=0.45\textwidth]{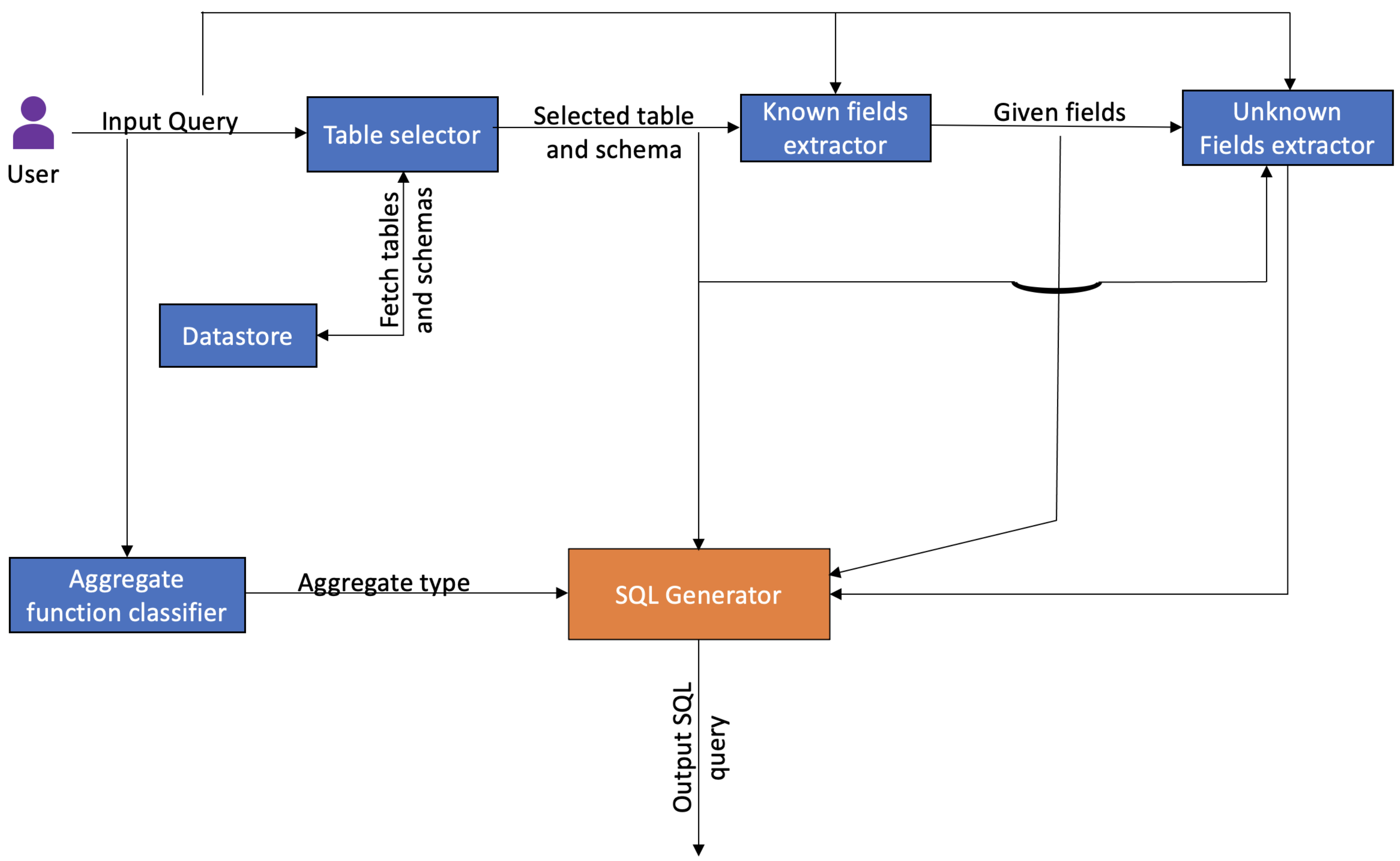}
  \caption{System Architecture of TableQuery} \label{fig:sysarch}
\end{figure*}

\begin{table*}[t]
\centering
\begin{tabular}{p{0.17\textwidth}p{0.17\textwidth}p{0.17\textwidth}p{0.17\textwidth}}
\hline
& \textbf{TableQuery} & \textbf{TAPAS} & \textbf{ln2sql}  \\
& \textbf{(Ours)} & \textbf{\citep{herzig-etal-2020-tapas}} & \textbf{\citep{couderc2015fr2sql}} \\
\hline
Data length & Any size & Max. sequence length 512* & Any size  \\
Debuggability & Yes & No & Yes  \\
Re-training & No$^{\plus}$ & Yes & No \\
Column types \\
\centering \small{Numeric} & Yes & Yes & Yes \\
\centering \small{String} & Yes & Yes & Yes \\
\centering \small{Date} & Yes & Yes & Yes \\
\centering \small{Categorical} & Yes & Yes & Yes \\
Supports Joins & No & No & Yes  \\
\hline
\end{tabular}
\caption{\label{tab:comparison}
\centering
Comparison of TableQuery with existing models to query on tabular data. \\
\small{* It is possible to run TAPAS for large datasets by splitting them into multiple small datasets of length 512. However, 1) Performance might suffer. 2) Multiple row queries will not work.} \\
\small{+ Please note that the question answering model in TableQuery can be switched for any other existing model trained on the free-text question answering dataset.}}
\end{table*}

\begin{table*}[!htp]
\centering
\begin{subtable}
\centering
\begin{tabular}{|l|l|l|l|l|l|l|}
\hline
\textbf{SNo} & \textbf{Year} & \textbf{Nationality} & \textbf{Gender} & \textbf{Cancer site} & \textbf{Death Count} & \textbf{age} \\
\hline
0&2016&Expatriate&Female&Liver And Intrahepatic Bile Ducts&1&50 \\
1&2013&Expatriate&Male&Stomach&1&55 \\
2&2017&Expatriate&Male&Oropharynx&1&55 \\
&&&&...&& \\
&&&&...&& \\
&&&&...&& \\
33&2013&Expatriate&Female&Colorectum&1&65 \\
34&2015&National&Female&Thyroid&1&50 \\
35&2015&National&Male&Leukaemia&1&75 \\
\hline
\end{tabular}
\label{tab:data}
\end{subtable}

\begin{subtable}
\centering
\begin{tabular}{p{0.9\linewidth}}
\\


\textbf{Give me the death count in 2012?} \\
\textbf{TAPAS:} SUM of 2, 3 \hfill \textcolor{teal}{\cmark} \\
\textbf{TableQuery:} SELECT SUM(death\_count) FROM cancer\_death WHERE year = '2012'  \hfill \textcolor{teal}{\cmark} \\
\vspace{0.05ex} 

\textbf{Give me death count of people below age 40 who had stomach cancer?} \\
\textbf{TAPAS:} SUM of 1 \hfill \textcolor{teal}{\cmark} \\
\textbf{TableQuery:} SELECT SUM(death\_count) FROM cancer\_death WHERE cancer\_site = 'Stomach' AND age  < 40  \hfill \textcolor{teal}{\cmark} \\
\vspace{0.05ex} 

\textbf{Give me death count between age 30 and 60 due to pancreas cancer?}
\\
\textbf{TAPAS:} 1 \hfill \textcolor{red}{\xmark} \\
\textbf{TableQuery:} SELECT SUM(death\_count) FROM cancer\_death WHERE cancer\_site = 'Pancreas' AND age  BETWEEN 30 AND 60   \hfill \textcolor{teal}{\cmark} \\
\vspace{0.05ex} 


\textbf{Get me the average deaths due to stomach cancer?} \\
\textbf{TAPAS:} AVERAGE of 1, 17, 30 \hfill \textcolor{red}{\xmark} \\
\textbf{TableQuery:} SELECT AVG(death\_count) FROM dataframe WHERE cancer\_site = \\ 'Stomach'\hfill \textcolor{teal}{\cmark} \\
\vspace{0.05ex}

\textbf{Get me the highest age for each cancer type?} \\
\textbf{TAPAS:} *No Output* \hfill \textcolor{red}{\xmark} \\
\textbf{TableQuery:} SELECT MAX(cancer\_site) FROM dataframe  \hfill \textcolor{red}{\xmark} \\
\vspace{0.05ex} 


\end{tabular}
\end{subtable}

\caption{ \label{tab:data}
\centering
Comparison of SOTA model, TAPAS and TableQuery (Ours). \newline
\textcolor{red}{\xmark} is an incorrect prediction. \textcolor{teal}{\cmark} is a correct prediction.
}
\end{table*}

This section discusses the system architecture of TableQuery as shown in the figure \ref{fig:sysarch}. Below we describe the main modules of TableQuery and how they are connected.

\textbf{Datastore: }
The datastore is a collection of tables that can be queried using natural language. These tables can be a database or a directory of spreadsheets. In addition to the tabular data, the datastore contains schema files that consist of metadata for each of the tables, such as table keywords, column names, column types, column keywords (user-provided keywords for column names), etc. The schema files can be created manually for these tables or automatically generated by applying various heuristics to the data. 

TableQuery supports various column types such as INTEGER, FLOAT, STRING, DATE, and CATEGORICAL. These types can have subtypes; for example, AGE and YEAR are INTEGER subtypes with appropriate ranges.
\\


\textbf{Table selector: }
The table selector selects the appropriate table from the datastore given an input query. This is done by extracting keywords from the input query and finding the table with the maximum overlap coefficient between the table's keywords from the schema (as well as column names, column keywords, etc.) and the question keywords. See appendix 2.1 for the pseudo-code of the table selector component.

\textbf{Known Fields Extractor: }
The known fields extractor extracts the columns for which values are already given in the query. The values corresponding to each of those columns are also extracted. This is done with the help of a deep learning model pre-trained to perform question answering on free text. For instance, consider the following user query on cancer death data in appendix table 1 - "How many men had stomach cancer in the year 2012?".  The known fields extractor outputs the column names as "gender," "cancer\_site," and "year," and their respective values as "male," "Stomach," and "2012". See appendix 2.2 for the pseudo-code of known fields extractor component.

\textbf{Unknown Fields Extractor: }
The unknown fields extractor extracts columns for which values have to be retrieved from the selected table. This is done by excluding columns for which values have already been extracted from the query by the known fields extractor
and finding the column with maximum overlap coefficient between its column keywords and the question keywords. The columns extracted using the known field extractor are excluded from the search. For instance, in the query "How many men had stomach cancer in the year 2012?" on the cancer death dataset in appendix table 1, the unknown field is "Death Count" See appendix 2.3 for the pseudo-code of unknown fields extractor component.

\textbf{Aggregate function classifier: }
While some queries require just the retrieval of values for unknown fields from the selected table, some might require performing further operations over those columns. This is achieved by including SQL aggregate functions such as COUNT, SUM, MIN, MAX, AVG, etc., in the generated SQL query. The aggregate function classifier decides the aggregate function to be used given an input query. To perform this task, we trained a two-layer neural network that takes the query encoded using Universal Sentence Encoder \cite{cer2018universal} as input and outputs the appropriate aggregate function to be used, if any. The model was trained on a commonly used dataset for tabular question-answering models, WikiSQL \cite{zhongSeq2SQL2017}, which was preprocessed to extract relevant components from it. We split our dataset into 80-20 training and testing splits and got an accuracy of 84.2 on the testing split. See appendix 2.4 for the pseudo-code of the aggregate function classifier component.

\textbf{SQL Generator: }
The SQL Generator combines the known fields and values, unknown fields, and aggregate function using the logic described in the pseudo-code (appendix section 2) to build an SQL query which, when run against the selected table, returns the desired result. See appendix 2.6 for the pseudo-code of the SQL generator component.






\begin{table*}[t]
\centering
\begin{subtable}
\centering
\begin{tabular}{|l|l|l|l|l|l|}
\hline
\textbf(Star (Pismis24-\#)) & \textbf(Spectral type) & \textbf(Magnitude (M bol )) & \textbf(Temperature (K)) & \textbf(Radius (R + )) & \textbf(Mass (M + )) \\
\hline
1NE & O3.5 If * & −10.0 & 42000 & 17 & 74 \\
1SW & O4 III & −9.8 & 41500 & 16 & 66 \\
2 & O5.5 V(f) & −8.9 & 40000 & 12 & 43 \\
3 & O8 V & −7.7 & 33400 & 9 & 25 \\
10 & O9 V & −7.2 & 31500 & 8 & 20 \\
12 & B1 V & −5.3 & 30000 & 4 & 11 \\
13 & O6.5 III((f)) & −8.6 & 35600 & 12 & 35 \\
15 & O8 V & −7.8 & 33400 & 10 & 25 \\
16 & O7.5 V & −9.0 & 34000 & 16 & 38 \\
17 & O3.5 III & −10.1 & 42700 & 17 & 78 \\
18 & B0.5 V & −6.4 & 30000 & 6 & 15 \\
\hline
\end{tabular}
\end{subtable}

\begin{subtable}
\centering
\begin{tabular}{p{0.9\linewidth}}
\\

\textbf{What is the smallest possible radius?} \\
\textbf{TAPAS:} SUM of 4 \hfill \textcolor{teal}{\cmark} \\
\textbf{TableQuery Answer:} [(4,)] \hfill \textcolor{teal}{\cmark} \\
\textbf{TableQuery:} SELECT MIN(Radius (R + )) FROM dataframe  \hfill \textcolor{teal}{\cmark} \\
\textbf{Expected Answer:} select MIN(Radius (R + )) FROM dataframe   \\
\vspace{0.05ex} 

\textbf{What are all the spectral types for star mismis24-\# is 1sw?} \\
\textbf{TAPAS:} O3.5 If *, O4 III \hfill \textcolor{red}{\xmark} \\
\textbf{TableQuery Answer:} [('1NE', 'O3.5 If *'), ('1SW', 'O4 III'), ('2', 'O5.5 V(f)'), ('3', 'O8 V'), ('10', 'O9 V'), ('12', 'B1 V'), ('13', 'O6.5 III((f))'), ('15', 'O8 V'), ('16', 'O7.5 V'), ('17', 'O3.5 III'), ('18', 'B0.5 V')]  \hfill \textcolor{red}{\xmark} \\
\textbf{TableQuery:} SELECT star\_pismis24\_, spectral\_type FROM dataframe  \hfill \textcolor{red}{\xmark} \\
\textbf{Expected Answer:} select spectral\_type FROM dataframe where star\_pismis24\_ = 1SW  \\
\vspace{0.05ex} 

\textbf{If a radius is 10, what  is the lowest possible mass?} \\
\textbf{TAPAS:} SUM of 11 \hfill \textcolor{red}{\xmark} \\
\textbf{TableQuery Answer:} [(25,)] \hfill \textcolor{teal}{\cmark} \\
\textbf{TableQuery:} SELECT MIN(mass\_m\_) FROM dataframe WHERE radius\_r\_ = '10'  \hfill \textcolor{teal}{\cmark} \\
\textbf{Expected Answer:} select MIN(mass\_m\_) FROM dataframe where radius\_r\_ = 10  \\
\vspace{0.05ex} 

\end{tabular}
\end{subtable}

\caption{ \label{tab:wikisql1}
\centering
Comparison of SOTA model, TAPAS and TableQuery (Ours) on WikiSQL dataset. (Table-id: 1-10015132-16) \newline
\textcolor{red}{\xmark} is an incorrect prediction. \textcolor{teal}{\cmark} is a correct prediction.
}
\end{table*}

\begin{table*}[b]
\centering
\begin{subtable}
\centering
\resizebox{\textwidth}{!}{%
\begin{tabular}{@{\extracolsep{\fill}}|l|l|l|l|l|l|l|l|l|}
\hline
\textbf{Season} & \textbf{Driver} & \textbf{Team} & \textbf{Engine} & \textbf{Poles} & \textbf{Wins} & \textbf{Podiums} & \textbf{Points} & \textbf{Margin of defeat} \\

\hline
1950 & Juan Manuel Fangio & Alfa Romeo & Alfa Romeo & 4 & 3 & 3 & 27 & 3\\
1951 & Alberto Ascari & Ferrari & Ferrari & 2 & 2 & 3 & 25 & 6\\
1952 & Giuseppe Farina & Ferrari & Ferrari & 2 & 0 & 4 & 24 & 12\\
1953 & Juan Manuel Fangio & Maserati & Maserati & 1 & 1 & 4 & 28 & 6.5\\
& & & .. & & & & & \\
& & & .. & & & & & \\
& & & .. & & & & & \\
2008 & Felipe Massa & Ferrari & Ferrari & 6 & 6 & 10 & 97 & 1\\
2009 & Sebastian Vettel & Red Bull & Renault & 4 & 4 & 8 & 84 & 11\\
2010 & Fernando Alonso & Ferrari & Ferrari & 2 & 5 & 10 & 252 & 4\\
2011 & Jenson Button & McLaren & Mercedes & 0 & 3 & 12 & 270 & 122 \\
\hline
\end{tabular}}
\end{subtable}

\begin{subtable}
\centering
\begin{tabular}{p{0.9\linewidth}}
\\

\textbf{Which podiums did the Williams team have with a margin of defeat of 2?} \\
\textbf{TableQuery Answer:} [('Renault', 7), ('Williams', 9), ('Ferrari', 9), ('McLaren', 10)] \hfill \textcolor{red}{\xmark} \\
\textbf{TableQuery:} SELECT team,podiums FROM dataframe WHERE margin\_of\_defeat = '2'  \hfill \textcolor{red}{\xmark} \\
\textbf{Expected Answer:} select MIN(Radius (R + )) FROM dataframe   \\
\vspace{0.05ex} 

\textbf{How many drivers on the williams team had a margin of defeat of 2?} \\
\textbf{TableQuery Answer:} [(1,)] \hfill \textcolor{teal}{\cmark} \\
\textbf{TableQuery:} SELECT COUNT(driver) FROM dataframe WHERE margin\_of\_defeat = '2' AND team = 'Williams'  \hfill \textcolor{teal}{\cmark} \\
\textbf{Expected Answer:} select COUNT(driver) FROM dataframe where team = Williams and margin\_of\_defeat = 2 \\
\vspace{0.05ex} 

\textbf{How many seasons was clay regazzoni the driver?} \\
\textbf{TableQuery Answer:} [(1,)] \hfill \textcolor{teal}{\cmark} \\
\textbf{TableQuery:} SELECT COUNT(season) FROM dataframe WHERE driver = 'Clay Regazzoni'  \hfill \textcolor{teal}{\cmark} \\
\textbf{Expected Answer:} select COUNT(season) FROM dataframe where driver = Clay Regazzoni   \\
\vspace{0.05ex} 

\textbf{Which margin of defeats had points of 30?} \\
\textbf{TableQuery Answer:} [('12',)] \hfill \textcolor{teal}{\cmark} \\
\textbf{TableQuery:} SELECT margin\_of\_defeat FROM dataframe WHERE points = '30'  \hfill \textcolor{teal}{\cmark} \\
\textbf{Expected Answer:} select margin\_of\_defeat FROM dataframe where points = 30 \\
\vspace{0.05ex} 

\textbf{Which podiums did the alfa romeo team have?} \\
\textbf{TableQuery Answer:} [(3,)] \hfill \textcolor{teal}{\cmark} \\
\textbf{TableQuery:} SELECT podiums FROM dataframe WHERE team = 'Alfa Romeo' \hfill \textcolor{teal}{\cmark} \\
\textbf{Expected Answer:} select podiums FROM dataframe where team = Alfa Romeo  \\

\end{tabular}
\end{subtable}

\caption{ \label{tab:wikisql2}
\centering
Performance of TableQuery (Ours) on WikiSQL dataset (Table-id: 1-10753917-1). TAPAS does not work for this example due to the length of the table. \newline
\textcolor{red}{\xmark} is an incorrect prediction. \textcolor{teal}{\cmark} is a correct prediction.
}
\end{table*}

\section{Results}
We compared the features of our tool with the existing state-of-the-art work in querying tabular data using natural language. Table \ref{tab:comparison} summarizes information about these models. TableQuery does not impose any data length restrictions, unlike TAPAS \citep{herzig-etal-2020-tapas} and allows to query data of arbitrary length. It does not require any re-training on the new dataset as the pre-trained model can be switched easily with any other model with better performance. Additionally, it is easy to debug mistakes in TableQuery compared to TAPAS as the final query is constructed piece-by-piece using the results of each module. 

Table \ref{tab:data} shows some examples comparing TableQuery and TAPAS. We performed this comparison on an open-source database available in the health category from Abu Dhabi Open Platform (\url{https://addata.gov.ae/}), an initiative to make datasets publicly available by the Government of Abu Dhabi. Due to the memory restrictions of TAPAS, we randomly filtered 36 rows from our existing cancer deaths dataset. TableQuery performs better than TAPAS when the queries are complex and multiple conditions are evaluated in the WHERE clause. Both the models perform equally well when the queries are relatively simple or a single condition is evaluated. The filtered 36 rows from the dataset used for testing are shown in appendix table 1.

\section{Ablation Study}
In this section, we share some qualitative results from experiments conducted on the WikiSQL dataset \cite{zhongSeq2SQL2017}, a dataset for developing language interfaces for relational databases. Since TableQuery currently does not support inner queries and joins, we do not perform a quantitative analysis; rather select a subset of tables without needing any joins. Tables \ref{tab:wikisql1} and \ref{tab:wikisql2} above highlight some of the selected results. TAPAS fails to generate results in cases where the table is large (example in \ref{tab:wikisql2}).

\section{Conclusion}
In this paper, we introduced TableQuery, a novel tool that allows querying data present in tabular format using natural language. Unlike other solutions, which are deep learning models that are trained end-to-end to perform this task, TableQuery uses existing question answering models pre-trained on free text. Our tool improves the limitations of existing work, such as the need to store the dataset into memory and re-training the model to expand the domain coverage.


\section*{Acknowledgements}
We want to thank the authors of HuggingFace transformers for making it easier to work with pre-trained deep learning models \cite{wolf-etal-2020-transformers} and the authors of Universal Sentence Encoder \cite{cer2018universal}. 

\bibliography{submission}
\bibliographystyle{acl_natbib}

\newpage

\appendix

\section{Appendix 1 - Cancer Death Dataset}

Table \ref{tab:abu-dataset} contains a random subset of cancer death dataset from the health data category available publicly on the Abu Dhabi Open Data Platform. 

\begin{table}[h]
\centering

\begin{tabular}{|l|l|l|l|l|l|l|}
\hline
\textbf{SNo} & \textbf{Year} & \textbf{Nationality} & \textbf{Gender} & \textbf{Cancer site} & \textbf{Death Count} & \textbf{age} \\
\hline
0&2016&Expatriate&Female&Liver And Intrahepatic Bile Ducts&1&50\\
1&2013&Expatriate&Male&Stomach&1&55\\
2&2017&Expatriate&Male&Oropharynx&1&55\\
3&2017&National&Male&Pancreas&2&50\\
4&2016&Expatriate&Male&Oropharynx&1&45\\
5&2012&Expatriate&Female&Pancreas&2&60\\
6&2014&National&Male&Leukaemia&2&70\\
7&2011&National&Male&Colorectum&2&75\\
8&2017&Expatriate&Female&Secondary Respiratory And Digestive Organs&1&65\\
9&2016&National&Female&Colon&1&40\\
10&2015&Expatriate&Female&Leukaemia&1&45\\
11&2018&National&Male&Rectosigmoid Junction&1&55\\
12&2018&Expatriate&Male&Colon&1&50\\
13&2014&National&Female&Liver and intrahepatic bile ducts&2&60\\
14&2014&Expatriate&Female&Trachea, bronchus and lung&2&60\\
15&2011&National&Male&Brain, nervous system&1&35\\
16&2016&National&Female&Bronchus And Lung&1&50\\
17&2016&National&Female&Stomach&1&20\\
18&2014&Expatriate&Male&Trachea, bronchus and lung&3&60\\
19&2018&Expatriate&Female&Breast&11&55\\
20&2016&Expatriate&Female&Pancreas&1&35\\
21&2014&Expatriate&Male&Leukaemia&5&45\\
22&2014&Expatriate&Female&Leukaemia&1&20\\
23&2012&Expatriate&Male&Pancreas&3&70\\
24&2018&National&Female&Bronchus And Lung&1&55\\
25&2011&National&Male&Lymphoma&1&60\\
26&2017&Expatriate&Male&Follicular [Nodular] Non-Hodgkin'S Lymphoma&1&55\\
27&2011&Expatriate&Female&Pancreas&1&60\\
28&2016&National&Female&Colon&1&85\\
29&2018&Expatriate&Male&Liver And Intrahepatic Bile Ducts&2&55\\
30&2011&National&Male&Stomach&1&70\\
31&2013&National&Male&Leukaemia&2&30\\
32&2014&Expatriate&Female&Other Cancer&1&60\\
33&2013&Expatriate&Female&Colorectum&1&65\\
34&2015&National&Female&Thyroid&1&50\\
35&2015&National&Male&Leukaemia&1&75\\
\hline
\end{tabular}
\caption{\label{tab:abu-dataset}
\centering
Random 36 rows of cancer death data from Abu Dhabi Open Data Platform (\url{https://addata.gov.ae/}). 
}
\end{table}

\section{Appendix 2 - Pseudo code}

Below we describe the pseudo-code which can be used to replicate the code for our tool.

\begin{algorithm}
\caption{ \textbf{2.1 Select Table}
}\label{select_tab}
\begin{algorithmic}
\Function{select\_table}{input\_query, datastore}
    \State $question\_tokens$ = \Call{tokenize}{$input\_query$} 
    \State $table$ = \Call{argmax}{$table$ $\in$ $datastore$} \Call{overlap\_coefficient}{$question\_tokens, table.keywords$}
    \State \textbf{return} $table$
\EndFunction
\end{algorithmic}
\end{algorithm}

\begin{algorithm}
\caption{ \textbf{2.2 Extract Known Fields}
}\label{extract_kwnfields}
\begin{algorithmic}
\Function{extract\_known\_fields}{input\_query, selected\_table}
    \State $known\_fields$ = \{\}

    \For{$col$ in $selected\_table.columns$}
        \If{\Call{is\_numeric}{$col.type$}}
            \State $col\_query$ = "how many " + $col$.name
        \Else
            \State $col\_query$ = "which are " + $col$.name
        \EndIf
        \State $column\_value$ = \Call{question\_answering}{$input\_query, col\_query$}
        \State $known\_fields[col.name]$ = \Call{adapt}{$column\_value,col.type$}
    \EndFor
    
    \State \textbf{return} $known\_fields$
\EndFunction
\end{algorithmic}
\end{algorithm}

\begin{algorithm}[h]
\caption{ \textbf{2.3 Extract Unknown Fields}
}\label{extract_unkfields}
\begin{algorithmic}
\Function{extract\_unknown\_fields}{input\_query, selected\_table, known\_fields}
    \State $unknown\_fields$ = []
    \State $question\_tokens$ = \Call{tokenize}{$input\_query$}
    \For{$col$ in $selected\_table.columns$}
        \If{$col$ not in $known\_fields$ and \Call{overlap\_coefficient}{$question\_tokens, col.keywords$} \textgreater THRESHOLD}
            \State Append $col.name$ to $unknown\_fields$
        \EndIf
    \EndFor
    \State \textbf{return} $unknown\_fields$
\EndFunction
\end{algorithmic}
\end{algorithm}

\begin{algorithm}[h]
\caption{ \textbf{2.4 Get Aggregate Outputs}
}\label{agg_op}
\begin{algorithmic}
\Function{get\_aggregate\_operator}{input\_query}
    \State $input\_query\_encoded$ = \Call{universal\_sentence\_encoder}{$input\_query$}
    \
    \State \textbf{return} \Call{aggregate\_operation\_classifier\_model.predict}{$input\_query\_encoded$}
\EndFunction
\end{algorithmic}
\end{algorithm}

\begin{algorithm}[h]
\caption{ \textbf{2.5 Get comparison output}
}\label{comp_op}
\begin{algorithmic}
\Function{get\_comparison\_operator}{column\_value}
    \State comparison\_operators = ["LESS THAN", "GREATER THAN", "MORE THAN", "BETWEEN", ...]
    \For{$comparison\_op$ in $comparison\_operators$}
        \If{$comparison\_op$ in value}
            \State \textbf{return} $column\_value.replace(comparison\_op, \Call{sql\_keyword\_for}{comparison\_op})$
        \EndIf
    \EndFor
    \State \textbf{return} $column\_value$
\EndFunction
\end{algorithmic}
\end{algorithm}

\begin{algorithm}
\caption{ \textbf{2.6 Generate SQL query}
}\label{comp_op}
\begin{algorithmic}
\Function{generate\_sql\_query}{input\_query,selected\_table, known\_fields,unknown\_fields,aggregate\_operator}
    \State $selected\_table$ = \Call{select\_table}{$input\_query$}
    \State $known\_fields$ = \Call{extract\_known\_fields}{$input\_query, selected\_table$}

    \For{$index, key, value$ in $0...length(known\_field)-1, known\_fields.keys, known\_fields.values$}
        \If{$index$ = 0}
            \State sql\_conditions = "WHERE" + $key$ + "IS" + $value$
        \ElsIf{$index >$ 0}
            \State sql\_conditions += "AND" + $key$ + "IS" + $value$ 
        \EndIf
    \EndFor
    
    \State $unknown\_fields$ = \Call{extract\_unknown\_fields}{$input\_query, selected\_table, known\_fields$}
    
    \State $aggregate\_operator$ = \Call{get\_aggregate\_operator}{$input\_query$}
    \State $output\_sql\_query$ = "SELECT" + aggregate\_operator + unknown\_fields + "FROM" + selected\_table.name + sql\_conditions
    
    \State \textbf{return} $output\_sql\_query$
\EndFunction
\end{algorithmic}
\end{algorithm}



\end{document}